\documentclass{vgtc}                          % final (conference style)
\ifpdf%                                % if we use pdflatex
  \pdfoutput=1\relax                   % create PDFs from pdfLaTeX
  \usepackage{graphicx}                % allow us to embed graphics files
  \DeclareGraphicsExtensions{.pdf,.png,.jpg,.jpeg} % for pdflatex we expect .pdf, .png, or .jpg files
\else%                                 % else we use pure latex
  \usepackage{graphicx}                % allow us to embed graphics files
  \DeclareGraphicsExtensions{.eps}     % for pure latex we expect eps files
\fi%

\graphicspath{{figures/}{pictures/}{images/}{./}} % where to search for the images

\usepackage{microtype}                 % use micro-typography (slightly more compact, better to read)
\PassOptionsToPackage{warn}{textcomp}  % to address font issues with \textrightarrow
\usepackage{textcomp}                  % use better special symbols
\usepackage{mathptmx}                  % use matching math font
\usepackage{times}                     % we use Times as the main font
\usepackage{cite}                      % needed to automatically sort the references
\usepackage{tabu}                      % only used for the table example
\usepackage{booktabs}                  % only used for the table example
\usepackage{amsmath,amssymb,amsfonts}
\usepackage{algorithmic}
\usepackage{textcomp}
\usepackage{xcolor}
\usepackage[textsize=footnotesize]{todonotes}
\usepackage{verbatim}
\usepackage[nolist]{acronym}
\usepackage{dirtytalk}
\usepackage{graphicx}
\usepackage{nicefrac}
\usepackage{comment}
\usepackage{algorithm} %Algorithm packages
\usepackage{enumitem}
\usepackage{multirow}
\usepackage{caption}
\usepackage{subcaption}
\usepackage{url}

\usepackage{morefloats}
\usepackage{soul}

 \newcommand{\change}[1]{\textcolor{black}{#1}}

\newacro{vr}[VR]{Virtual Reality}
\newacro{hmd}[HMD]{Head-Mounted Display}
\newacro{ems}[EMS]{Electrical Muscle Stimulation}
\newacro{imu}[IMU]{Inertial Measurement Unit}
\newacro{bci}[BCI]{Brain-Computer Interface}
\newacro{vwg}[VWG]{Virtual World Generator}
\newacro{moo}[MOO]{Multi-Objective Optimization}
\newacro{dof}[DOF]{Degrees of Freedom}
\newacro{ssq}[SSQ]{Simulator Sickness Questionnaire}
\newacro{fov}[FOV]{Field of View}
\newacro{plt}[PLT]{Pareto Least Turns}
\newacro{psp}[PSP]{Pareto Shortest Path}
\newacro{rrt}[RRT]{Rapidly-exploring Random Trees}
\newacro{vims}[VIMS]{Visually Induced Motion Sickness}
\newacro{ddr}[DDR]{Differential Drive Robot}
\newacro{vrise}[VRISE]{VR Induced Symptoms and Effects}
\newacro{osf}[OSF]{Open Science Foundation}
\newacro{wb}[WB]{Wii Balance Board control method}
\newacro{js}[JS]{Joystick control method}
\newacro{ve}[VE]{Virtual Environment}
\newacro{wii}[Wiiboard]{Wii Balance Board}
\newacro{tlx}[NASA-TLX]{NASA Task Load Index}

\onlineid{0}

\vgtccategory{Research}

\title{%A comfortable sickness: 
Leaning-Based Control of an Immersive-Telepresence Robot}

\author{Joona Halkola, Markku Suomalainen, Basak Sakcak, Katherine J. Mimnaugh,\\ Juho Kalliokoski, Alexis P. Chambers, Timo Ojala and Steven M. LaValle \\ \textit{Center for Ubiquitous Computing, Faculty of Information Technology and Electrical Engineering}, \\ \textit{University of Oulu,} Oulu, Finland\thanks{email: firstname.lastname@oulu.fi} }
\abstract{In this paper, we present an implementation of a leaning-based control of a differential drive telepresence robot and a user study in simulation, \change{with the goal of bringing the same functionality to a real telepresence robot}. The participants used a balance board to control the robot and viewed the virtual environment through a head-mounted display.
The main motivation for using a balance board as the control device stems from Virtual Reality (VR) sickness; even small movements of your own body matching the motions seen on the screen decrease the sensory conflict between vision and vestibular organs, which lies at the heart of most theories regarding the onset of VR sickness.
To test the hypothesis that the balance board as a control method would be less sickening than using joysticks, we designed a user study (N=32, 15 women) in which the participants drove a simulated differential drive robot in a virtual environment with either a Nintendo Wii Balance Board or joysticks. However, our pre-registered main hypotheses were not supported; the joystick did not cause any more VR sickness on the participants than the balance board, and the board proved to be statistically significantly more difficult to use, both subjectively and objectively. Analyzing the open-ended questions revealed these results to be likely connected, meaning that the difficulty of use seemed to affect sickness; even unlimited training time before the test did not make the use as easy as the familiar joystick. Thus, making the board easier to use is a key to enable its potential; we present a few possibilities towards this goal.
} % end of abstract

\CCScatlist{
  \CCScatTwelve{Telepresence robot}  {VR sickness} {Balance board} {User comfort}
}

\begin{document}

%% The ``\maketitle'' command must be the first command after the
%% ``\begin{document}'' command. It prepares and prints the title block.

%% the only exception to this rule is the \firstsection command

\maketitle

\section{Introduction}\label{sec:intro}
Immersive telepresence, the ability to be and \textit{feel} present in a remote location through technology, has the potential to enable hybrid meetings (including both remote and local participants) where the remote attendees feel like they really were there; it has been shown that with current regular screen-based telepresence robots, the remote users operating the robot speak less and consider group work more difficult \cite{stoll2018wait}. A probable reason for this 
%unnatural behavior of the 
type of (unnatural) behavior observed in
remote participants is the lack of \textit{presence} \cite{slater1997framework}, the feeling of "being there", which in virtual environments (VEs) makes users behave more naturally \cite{slater2009place}. It has been shown that regular, non-immersive screens do not make users feel as present as an \ac{hmd} \cite{slater2018immersion}. \change{Thus, further research on users embodying a physical robot is warranted.}

Besides a more immersive screen, another piece shown to increase the feeling of presence is the ability to embody a robot that one can move, instead of a stationary camera \cite{rae2014bodies}. However, immersive telepresence needs more contemplation than simply attaching a 360\textdegree\, camera to a mobile robot to drive around. The user must be able to control the robot with little effort but without succumbing to VR sickness \cite{laviola2000discussion}. Whereas there is a vast library of research on locomotion in VR, the case of travelling aboard a moving ground vehicle is often avoided; this kind of motion, especially rotations, has been shown to significantly contribute to VR sickness \cite{hu1999systematic,kemeny2017} and is often replaced with teleportation in virtual environments. However, when controlling physical systems, teleportation is not an option and more research is needed.

\begin{figure}
    \centering
    \begin{subfigure}[b]{0.3\textwidth}
        \includegraphics[trim=5 0 220 0,clip,width=0.95\textwidth]{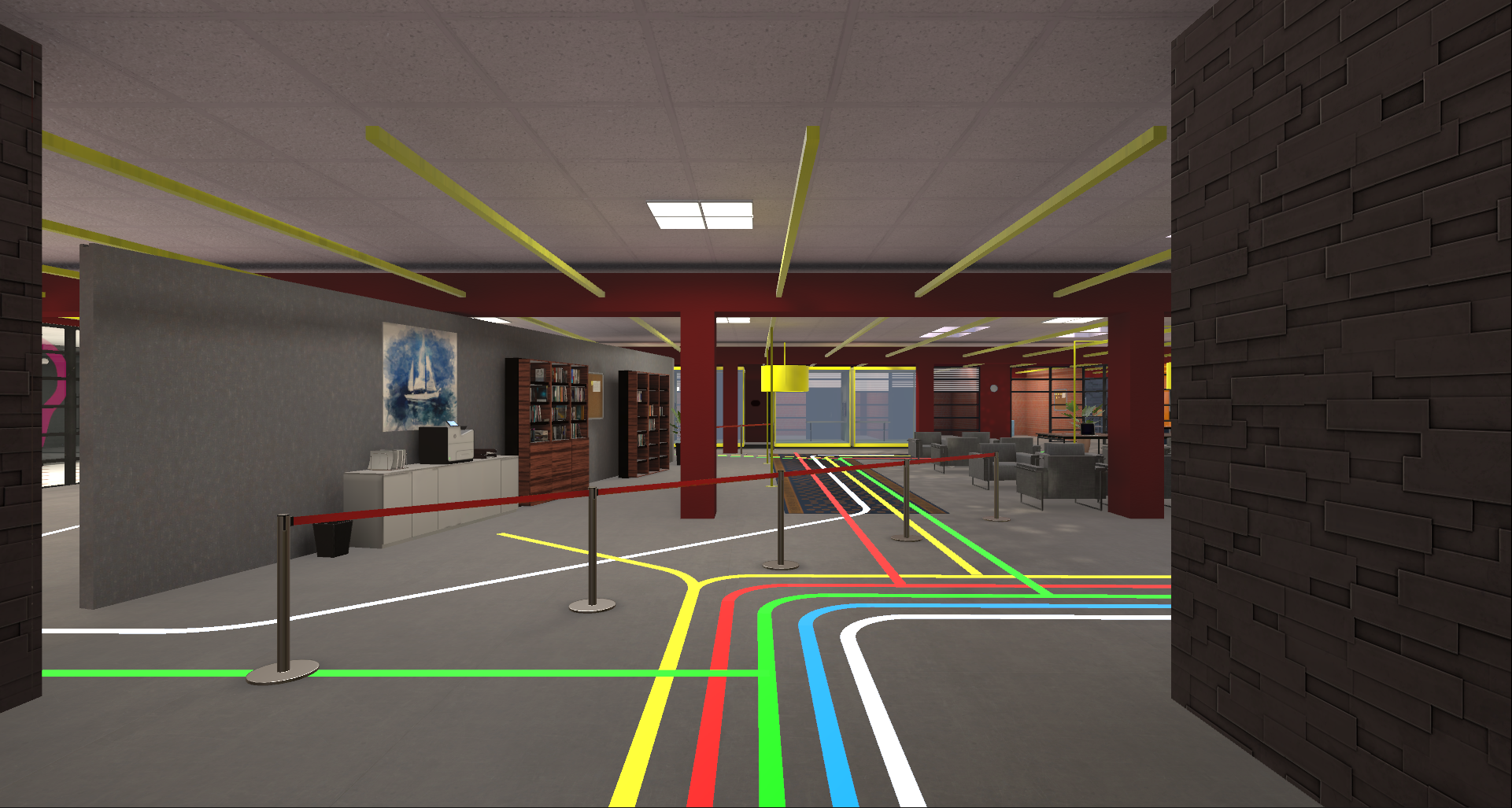} \vspace{-2mm}
        \caption{}
        \label{fig:start}
        \includegraphics[trim=15 30 10 30,clip,width=0.95\textwidth]{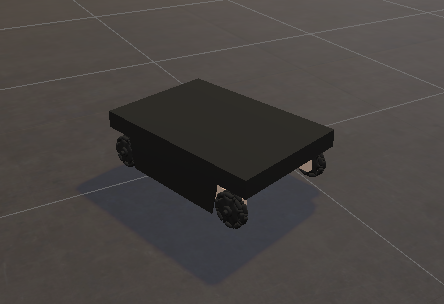}\vspace{-2mm}
        \caption{}
        \label{fig:robot}
    \end{subfigure}%
    \begin{subfigure}[b]{0.2\textwidth}
        \centering
        \includegraphics[width=0.97\textwidth]{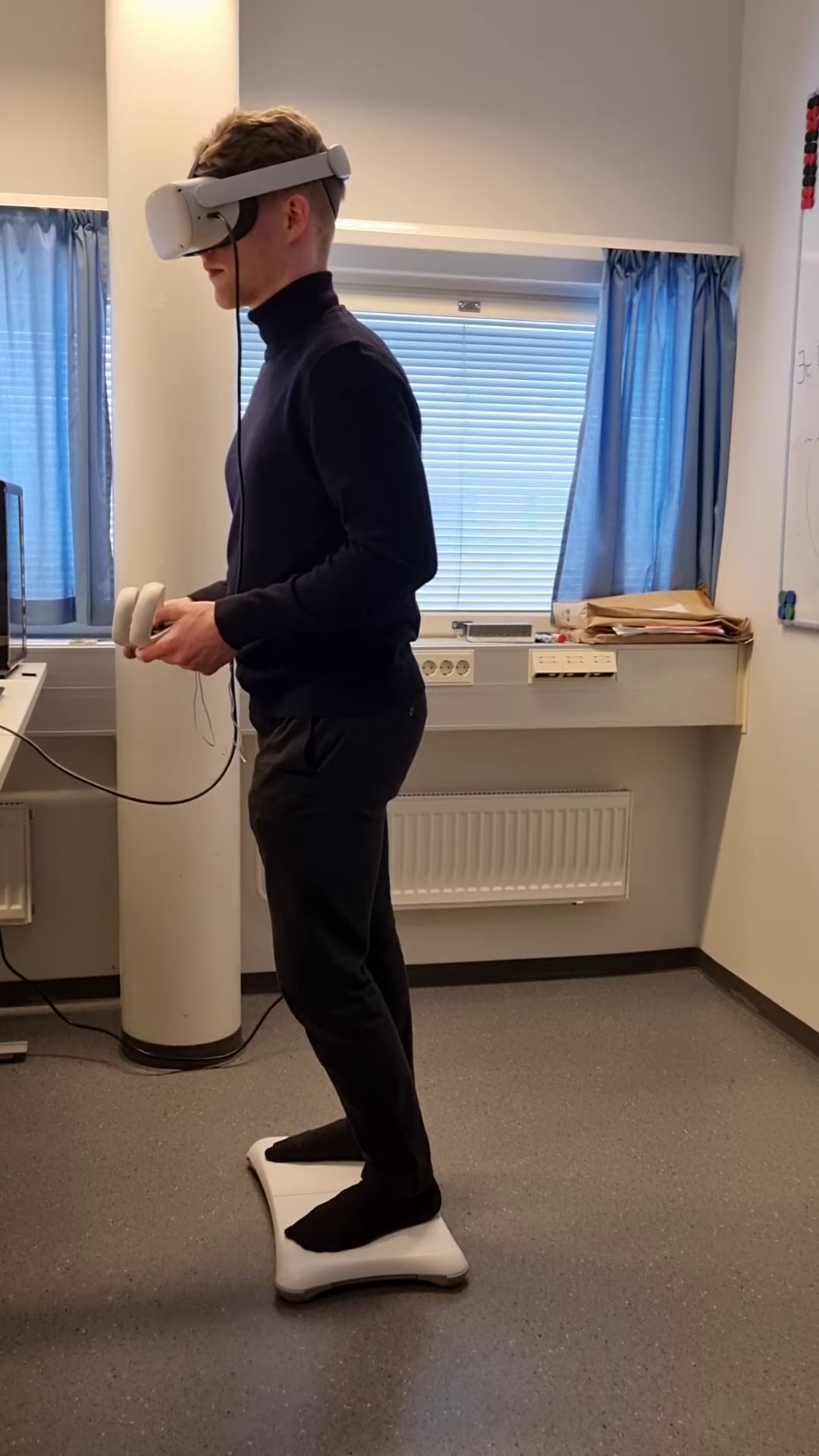}
        \vspace{-1mm}
        \caption{}
        \label{fig:teleop}
    \end{subfigure}%
\caption{(a) Virtual environment used in the user study. (b) The model of the telepresence robot's base, into which the pole with a camera and other equipment is attached. (c) An experimenter standing on a Wiiboard for leaning-based control of a telepresence robot, he is slightly leaning forward.}
\end{figure}

In this paper we explore the use of a balance board, particularly the Nintendo \ac{wii}, for leaning-based control of a telepresence robot while being immersed in an \ac{hmd}. The study is done as a simulation in a \ac{ve} using an accurate virtual model of the mobile robot at the research group's disposal (Probot Dolly differential drive robot, see Fig.\ref{fig:robot}). Even though we acknowledged that participants will be unfamiliar with the \ac{wii} and it might provide various physical challenges, we hypothesized that the use of body to control the robot would make the experience less sickening than the joystick due to the ability to mitigate sensory mismatches. However, it turns out that we underestimated the difficulty of using the balance board, partly based on earlier work which reports using it as easy \cite{de2008using}. 
In our study, the joystick was found statistically significantly easier to use than the balance board, both subjectively and objectively. 
Furthermore, we found that the leaning-based control method using \ac{wii} induced more VR sickness; 
%there was no significant difference in perceived VR sickness between the control conditions; 
open-ended answers reveal that difficulty of the \ac{wii} may have also played a role in the sickness results. We also discuss the findings about presence across control methods.

The novelty in this paper is the development of a control mapping to drive a telepresence robot using a balance board, and a rigorous user study done in a \ac{ve} using a virtual model of a real telepresence robot. Even though the results were not in favor of using this particular board in the way we imagined, the user study results show the potential of the method with better hardware and more training time than what was allowed in this study.

\section{Related work} \label{sec:related}
There is a rising interest in HMD-based immersive telepresence, due to its potential in enabling hybrid meetings. The motivation for HMD-based telepresence instead of using current commercial telepresence robots, such as the Double and Gobe, stems from the observation that participants through such telepresence robots are much less engaged in, for example, group work \cite{stoll2018wait}. This is likely related to \textit{presence}, a well researched concept in VR showing that the immersion of an HMD gives you a better sense of "being there" than a regular screen \cite{slater2018immersion}. Thus, there is increasing interest in immersive telepresence: recent use cases include, for example, education \cite{botev2021immersive}, earthquake rescue \cite{negrello2018humanoids} and social interaction \cite{du2020human}.

There is a wide variety of different locomotion methods for virtual environments, with well established advantages and disadvantages; for a comprehensive survey see, e.g., \cite{al2018virtual}. With a scalable method also usable in smaller spaces in mind when considering choices for HMD-based control of a telepresence robot, we decided not to consider walking (redirected or regular) but instead opted for leaning-based methods, keeping in mind the popular Segway robots, which the leaning-based methods mimic (even though the experience without actual motions is quite different).
There are various methods for implementing leaning-based navigation for VR; by harnessing the user, methods such as the human joystick \cite{marchal2011joyman} can allow leaning that properly resembles falling. Another popular method is simply tracking the HMD to detect leaning, or stepping, with different tactile feedback from the ground to help the user \cite{chen20136dof,nguyen2019naviboard}, which has been shown to increase presence and decrease VR sickness over joystick; similar methods have also been shown to work while sitting \cite{hashemian2021leaning}.

Of the works using \ac{wii}, de Haan et al.~\cite{de2008using} described the technical methods and positive initial user feedback for using Wiiboard for locomotion in VR. Valkoc et al.~\cite{valkov2010traveling} then used Wiiboard as a component in their navigation system, however, they did not compare Wiiboard to another control method. Thus, especially with the symmetries between Wiiboard, Segway, and \ac{ddr} (rotation is performed with different speeds on each wheel, which can be controlled with the pressure on each foot), we believe it is necessary to properly test a balance board in this use; comparing with the stepping motion, we hypothesize that small adjustments to the robot's direction would be easier to do with just weight adjustment on the board when compared to trying to make the same adjustment in the head-tracking with the whole body, as in \cite{chen20136dof,nguyen2019naviboard}. However, there is little previous work on VR sickness using a balance board, for either 
virtual environments or immersive telepresence; regardless of the vast literature on VR sickness in virtual environments (see, e.g.,~\cite{rebenitsch2016review}), in such ground-based motions VR sickness is often mitigated by teleportation, which is infeasible for telepresence. For autonomous motions of a telepresence robot with an HMD, several considerations were given to VR sickness in \cite{becerra2020human,suomalainen2022unwinding}; however, there was no consideration on the actual control of the robot, since users often want to control the robot also manually, especially for short distances. 

In using a balance board for controlling a robot, there are both potential mitigating and escalating factors for VR sickness. From one point of view, having to move yourself may mitigate sensory mismatch between the visual system and the vestibular organs, which should prevent VR sickness. However, another issue to consider is the \textit{postural stability}, deterioration of which has been shown to have the potential to predict the onset of VR sickness \cite{litleskare2021relationship}, even with Wiiboard as the measurement device when watching a movie through the HMD \cite{bos2013cinerama}. However, the order of cause and consequence is still complex \cite{litleskare2021relationship}; thus, it is difficult to predict whether having to balance on purpose can have a causal effect on VR sickness, and whether this has a stronger effect on VR sickness than the decrease of the sensory mismatch.

\section{System}\label{sec:system}

The system comprised a \ac{ve} (see Sec.~\ref{sec:env_and_taks} for a more detailed description) including a simulated robot (Fig.\ref{fig:robot}) moving within the environment
%which is viewed by the participants through an \ac{hmd}. 
\change{A virtual 360\textdegree~camera is attached 1.5 meters above the robot base, from which the user sees the virtual world through an \ac{hmd}; this is a height suggested by \cite{keskinen2019effect} for 360\textdegree~videos.}
\change{The position of the camera is fixed with respect to the robot while allowing 3 \ac{dof} in rotation as determined by the headset tracking. The robot can translate and rotate in a two-dimensional plane determined by the user input. In total, the robotic system has 5\ac{dof} since the camera height is fixed.}
The robot motion was controlled directly by the users through the inputs given using a \ac{wii} or the joysticks on the Oculus Quest 2 controllers. 
The user input was mapped to reference angular velocities for the robot wheels. The maximum attainable wheel angular velocity was set to approximately $573^\circ$ per second to limit the robot forward speed to a maximum of $0.75 m/s$. The reason for this choice was motivated by \cite{mimnaugh2021analysis} presenting that $1m/s$ is a suitable speed; however, we used a slightly lower value compared to precedents to increase comfort in controlling the robot using a device (\ac{wii}) that most participants would be unfamiliar with. 

\subsection{\ac{wii} control method}
\begin{figure}[ht]
\centering
\includegraphics[width=0.5\columnwidth]{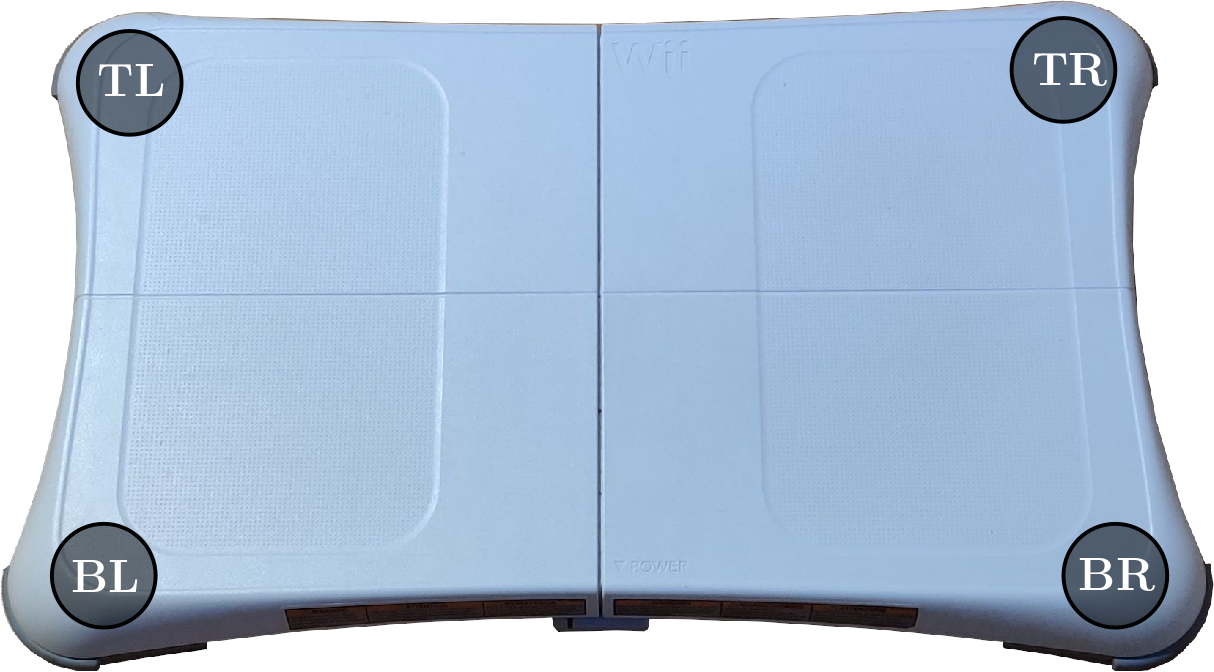}
\caption{\change{The Nintendo Wii Balance Board with the four force sensors at the corners marked.} }  %The videos used and other relevant documents are available at: \textcolor{red}{[redacted for blind review]}
\label{fig:wiiboard} 
%\vspace{-0.2cm}
\end{figure}

The \ac{wii} has four pressure sensors, one at each corner of the board, and communicates with a computer via Bluetooth (see Fig.~\ref{fig:wiiboard}). 
We used the WiiBuddy Unity Asset of BitLegit \cite{Wiibuddy} to establish the connection between the \ac{wii} and the PC together with \cite{Wiimote} to stabilize the connection.

Forward movement with \ac{wii} was achieved by leaning forward, and turning while moving forward by leaning towards either of the front corners. \change{The initial implementation followed the guidelines of \cite{de2008using}. However, due to the observed difficulty during initial tests of backward motion (by putting pressure on both heels) and rotation in place (by putting pressure on the heel with one foot and toe on the other), we decided for a simpler implementation, with further motivation being the low probability of needing backward motion for a telepresence robot. Thus, we implemented rotation in place by} putting weight towards one of the back corners, which was reasonable to achieve by turning the upper body towards the desired corner, similarly to reaching for an object behind you.

The sensor readings, corresponding to the applied pressure on each corner, were used to calculate the two components that constitute the total motion: go forward ($fwd$) and turn ($turn$). These were calculated according to the equations, \change{in which $F_\cdot$ is the force measured by the sensor at the respective corner,} (see Fig.~\ref{fig:wiiboard} for the meaning of each index term)
\begin{equation}
\begin{alignedat}{2}
fwd &= \max\left(\frac{(F_{TL}+F_{TR})-(F_{BL}+F_{BR})}{(F_{TL}+F_{TR})+(F_{BL}+F_{BR})},0\right)\\
turn &= \frac{F_{TR}-F_{TL}}{F_{TR}+F_{TL}}.
\end{alignedat}
\label{eq:fwd_turn}
\end{equation}
Note that $fwd$ and $turn$ lie within the ranges $[0,1]$ and $[-1,1]$, respectively. 
In case of a collision with the environment, the forward motion was disabled and the user was expected to rotate in place until the robot was free to move forward. In that case, the $turn$ component was calculated as 
\begin{equation}
    turn = \frac{F_{BR}-F_{BL}}{F_{BR}+F_{BL}}
    \label{eq:back_turn}
\end{equation}
and $fwd$ was set to $0$. Respective reference wheel rotational velocities were calculated as 
\begin{equation}
    \omega_L= c_ffwd + c_t\,turn, \; \omega_R=c_f fwd - c_t\,turn
    \label{eq:wheel_speed}
\end{equation}
in which $\omega_L$ and $\omega_R$ are the reference velocities for the left and right wheels, %and $\omega_{max}=573 deg/s$.
and $c_f$ and $c_t$ are the respective weights. The weights were selected considering limits on forward and turning speeds and ensuring that $\omega_L, \omega_R \leq \omega_{max}$. In case of rotate in place motion, used for getting out of a collision state, $\omega_{max}=75 deg/s$ was used to avoid uncomfortably fast rotations. This value was selected among three candidates ($132 deg/s$, $103 deg/s$, $75 deg/s$) by testing it on six people. 

%%%Calibration
For each user, the \ac{wii} was calibrated to account for varying weight distribution patterns during standing still and leaning. This way, the mapping from sensor readings to robot controls was made to adjust Wiiboard for each user individually. 
%To address the challenge of having a variety of people with different weights and leaning techniques, a calibration process for the sensor mapping from sensed weights to robot controls was made to adjust Wiiboard for each user individually. 
In the calibration process, values from the sensors are measured for six different cases: standing still, leaning forward, and leaning towards each of the four corners. Measurements corresponding to standing still are used as offsets to ensure that the robot does not move at this posture since $fwd$ and $turn$ are zero only when the weight is equally distributed (see Eqs~\eqref{eq:fwd_turn} and \eqref{eq:back_turn}). The rest of the cases were used to find maximum values achieved for $fwd$ and $turn$ when the participants leaned \emph{as much as they felt comfortable}. These values are then mapped to the $[-1,1]$ interval so that when the participant was at an extreme posture the robot speed was the highest.  
Since most users leaned too much when calibrating (even when told to ``lean as much as you feel comfortable"), sensor readings corresponding to maximum speeds were calculated by taking 60\% of the calibration values.

\subsection{Joystick control method}
Two joysticks on the Oculus Quest 2 controllers were used as an alternative to contrast the \ac{wii} control method. Whereas one joystick was used to control the forward speed ($fwd$), the other one was used to control the turning speed and direction ($turn$). The joystick position on the horizontal axis (moving the joystick left-right) was mapped to intervals $[0,1]$ and $[-1,1]$ corresponding to $fwd$ and $turn$, respectively. Robot motion is achieved by tracking the reference wheel rotational speeds, calculated in terms of $fwd$ and $turn$ using Eq.\eqref{eq:wheel_speed}. \change{Participants were standing still when using the joystick in the study}.

\section{Hypotheses}
We pre-registered the following three hypotheses, together with the procedure and analyses to be used in the study, in \ac{osf}\footnote{ \url{https://osf.io/6rxby/?view_only=9990df8e0c104ac9b4bbc03ac72515a8}}. From now on we will refer to the \textbf{\ac{wb} condition as WB} and \textbf{\ac{js} condition as JS}.

\begin{enumerate}[label=\textbf{H\arabic*:}]
    \item Less VR sickness in WB condition as indicated by lower total weighted \ac{ssq} score. 
    \item No difference in how far the participants get during the four minutes driving time and no difference in the number of crash events.
    \item No difference in cognitive load as indicated by no difference in NASA-TLX scores in any dimension except the physical load. 

\end{enumerate}

\change{We hypothesized H1 based on the evidence that incorporating some form of vestibular stimulation can potentially mitigate sensory conflict in VR, as described in Section~\ref{sec:related}. Even though there is a  slight mismatch between the vestibular and visual stimuli (such as forward-leaning being circular and not linear motion, and the rotations providing only an impulse of the motion that then further carries on visually in VR while the vestibular stimulation from the bodily motion ends), we believe that some amount of vestibular stimuli should ameliorate VR sickness as even random stimulation of the vestibular organs has been shown to help \cite{weech2018influence}. H2 stems from the work of de Haan el al. \cite{de2008using}, who reported subjects found the Wiiboard to be easy to use; however, due to our expectation that our sample would have some familiarity with using a joystick, we predicted that there would not be any difference in performance. We postulated H3 for similar reasons; the Wiiboard clearly needs more physical effort, but otherwise, based on the ease of use reported in \cite{de2008using}, we expected there would be no difference in cognitive load}.

\section{User Study}
\subsection{Procedure}

Two conditions with two different starting positions (either ends of the path shown in Fig.~\ref{fig:path}) were tried by the participants in a counterbalanced order such that each pair of starting position and control method was experienced as first and second equal number of times. 
The participant was welcomed by a researcher upon arrival and asked to sign a consent form. To pre-screen already sick-feeling people, the participants were asked if they felt nauseous or had a headache, and asked to reschedule if they did. After this, they were instructed to take their shoes off, stand on the Wiiboard, and place their feet in the middle of the textured areas of the Wiiboard (see Fig.~\ref{fig:wiiboard}). 

Next, the researcher gave instructions about the practice session to the participant; %which depended on given the control method. I
if the control method was the Wiiboard, the participant was first shown an instruction video on how to calibrate the balance board, after which the researcher instructed the participant how to put on the HMD and told the participant to complete the calibration. Next, the participant was told to take off the HMD and a second instruction video was shown to the participant which explained how to use the Wiiboard; this was done because the calibration was better to do with the HMD on to get realistic values to use with the HMD, but practicing with the HMD on would likely have caused too much VR sickness on the participants. If the control method was the joysticks, the researcher read out instructions on how to use the joysticks to move the robot. Finally, the researcher read out the instructions for the practice session, which were identical for both control methods: the participant was told to complete the practice session without the HMD and follow the robot movement from a monitor in front of them. The participant was specifically instructed to crash into a wall to test out the turn-in-place feature and learn how to 
recover from a collision.
%get unstuck from the collision avoidance system. 
After the practice, the participant was asked how confident they felt using the control method on a scale from 1 to 7 and the researcher marked down the answer. Then, the participant was told to put back on the HMD, follow the direction of the white line on the ground and count the sailboat paintings on the walls.

After the participant completed the run, they were told to take off the HMD and controllers and fill out questionnaires on a laptop, after which the same procedure was repeated for the second control method. At the end, the participant was rewarded with a 12€ gift voucher to Amazon and given a short debrief about the study.

\subsection{Virtual environment and tasks}\label{sec:env_and_taks}
\begin{figure}
\centering
\includegraphics[width=0.95\columnwidth]{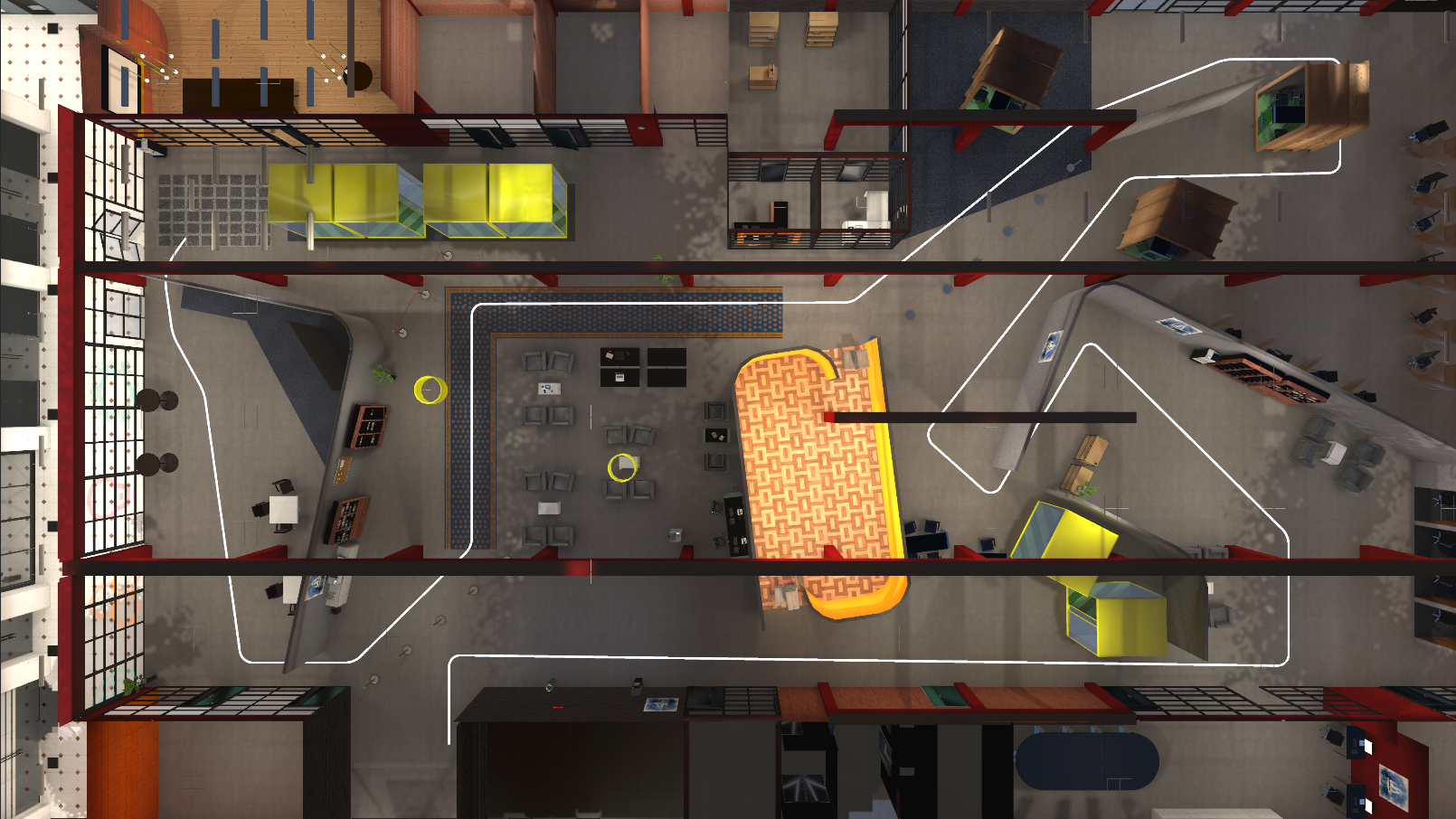}
\caption{Birds-eye view of the path (white line) the participants should follow in the main task. The other lines have been removed for clarity.}  %The videos used and other relevant documents are available at: \textcolor{red}{[redacted for blind review]}
\label{fig:path} 
%\vspace{-0.2cm}
\end{figure}

\ac{ve} for the study was loosely based on parts of the local university; the \ac{ve} and the path in the environment that the participants were asked to follow can be seen in Fig.~\ref{fig:path}. Each participant ran into opposite direction on the second attempt (combinations of directions and conditions were also counterbalanced). 
The environment had five differently colored lines on the floor as shown in Fig.~\ref{fig:start}, from which the participants were told to follow the white line during the test. This served as a sidetask to test whether the participants could still pay attention to the environment while commanding the robot. Another sidetask was the counting of six sailboat paintings scattered around the \ac{ve}. The training area was an empty hall with a line drawn on the floor to work as reference.

A basic collision-avoidance method is implemented that stops the robot if it is too close to an obstacle (a wall or an object). In case of such an event, the forward motion was disabled and only turn-in-place motion was allowed, until the robot was heading sufficiently away from the obstacle.
%A collision will halt any movement and prevent further forward movement, allowing only rotate-in-place functionality to be used to rotate the robot away from the collided object, after which forward movement is enabled again. %This way the subject is faced away from the object to prevent further collisions with the same object. 
The system measures the total amount of time spent being stuck in collision states.% and register if the subject has been facing the wall over 30 seconds which is used as one of the exclusion criteria.

\subsection{Participants}
Sample size for the study was 32 participants and we aimed at having an equal split of male and female participants. However, due to difficulties of getting enough female participants we ended up with 17 male and 15 female participants. All participants reported having normal or corrected-to-normal vision and none of the participants were colorblind.

VR system usage reported by the participants was: $12.5\%$ never, $40.6\%$ once or just a couple of times, $25.0\%$ once or twice a year, $15.6\%$ once or twice a month, $3.1\%$ once or twice a week, and $3.1\%$ every day. Responses to how often they play computer games were: $12.5\%$ never, $15.6\%$ once or just a couple of times, $9.4\%$ once or twice a year, $18.8\%$ once or twice a month, $21.9\%$ once or twice a week, $12.5\%$ several times a week, and $9.4\%$ every day.

Initially, we pre-registered that we would exclude all participants who got stuck against the walls for more than 30 seconds; the time limit was based on the same initial test that we used to choose the rotation speed. However, the number of participants who exceeded this limit were higher than what we expected. Due to the difficulties in recruiting participants, we decided to relax this limit and used also the data corresponding to the participants who were stuck for more than $30$ seconds but otherwise completed the required tasks (4 males, 3 females among 32 participants).
%However, due to the difficulties of getting enough participants and a higher amount of participants exceeding this limit as expected (4 males, 3 females), we had to relax this limit; 
The pre-registered hypotheses were run both with all participants ($N=32$) and the subset who did not get stuck for more than $30s$ ($N=25$). On average, participants were stuck for $21.05s$ in WB condition and $0.87s$ in JS condition for N=32; averages were $7.39s$ in WB condition and $0.32s$ in JS condition for N=25. Each session lasted exactly $4$ minutes.

\subsection{Measures}
Two sets of questionnaires were presented to the participant after finishing each of the sessions. The latter part of the second questionnaire contained post-experiment questions. Both questionnaires started with \ac{ssq} \cite{kennedy1993simulator} and \ac{ssq} total score was used to measure sickness. \ac{ssq} is an established questionnaire for measuring sickness in VR by presenting 16 possible sickness symptoms, which the participants gauge on a scale none (0) to severe (3). The SSQ total score is calculated by weighting the answers for a maximum score of 236. 
Higher scores indicate greater levels of sickness experienced.
Consequently, \ac{tlx} questionnaire \cite{TLX} was administered, which is used to measure six dimensions of workload (mental demand, physical demand, temporal demand, performance, effort, and frustration) of the task. Each dimension was rated on a scale of 1 to 20 and higher values indicate that the task is more demanding in that aspect. Additionally, we used 7-point Likert-scale questions, forced-choice questions comparing the two methods, and open-ended questions about reasons for some choices and demographic questions.

%Explain SUS if we add the results.. 

To measure the participants' confidence in using each control method we asked how confident the participant felt (on a 7-point Likert-scale) after each training session about the condition that they would try next. Their answer was then registered by the experimenter. 

For each session, the path executed by the robot was recorded in world coordinates and how far along the reference path (see Fig.~\ref{fig:path} for the reference path) the participant got was calculated projecting the final robot position onto the reference path. We also recorded the number of times the robot collided with the obstacles (including walls) and the total length of time spent in a collision-state.

\section{Results}
All statistical tests were run in SPSS with significance levels set to $0.05$ and with a $95 \%$ confidence interval.

\subsection{Confirmatory results}
All confirmatory analyses were performed first using the whole dataset $(N=32)$ and then using only the data corresponding to the participants who were not in a collision-state for longer than $30s$.

%\paragraph{\textbf{Less VR sickness in \ac{js} condition}}
\textbf{Less VR sickness in \ac{js} condition (H1 rejected)} \hspace{0.5em}
A Wilcoxon Signed-Ranks test (two sided) was performed to compare the differences between the total weighted SSQ scores for \ac{js} (${Mean}=32.02$) and \ac{wb} (${Mean}=40.44$) conditions $(N=32)$. The test indicated that \ac{js} elicited significantly lower SSQ scores compared to \ac{wb}, $Z=-2.32$, $p=.02$, $r=0.41$. We also checked for potential effect of getting stuck for more than $30s$ on sickness and run the test again using only the participants who did not get stuck or got stuck less than $30s$ $(N=25)$. We did not observe any significant difference between the total weighted SSQ scores for \ac{js} $(Mean=29.92)$ and \ac{wb} $(Mean=37.84)$ conditions when participants who got stuck were excluded from the analysis, as indicated by a Wilcoxon Signed-Ranks test (two sided), $Z=-1.84$, $p=.066$, $r=0.369$. 

\vspace{0.2em}
\noindent\textbf{Participants reached farther along the path in \ac{js} condition (H2 rejected)}\hspace{0.5em} A paired t-test was run to determine whether in one condition the participants reached farther along the path $(N=32)$. The total path length was $186.6$ meters and none of the participants reached the end (we ensured this by selecting the path length and the time limit on each session so that all participants have equal exposure to the \ac{ve}). The distances in both conditions were normally distributed, as indicated by a Shapiro-Wilk test, $W(32)=0.956, p=.214$ for \ac{js}, $W(32)=0.966$, $p=.398$ for $\ac{wb}$, and there were no outliers in the data, as assessed by inspection of the boxplots. The mean distance was higher in \ac{js} condition $(162.39 \pm 13.52)$ compared to in \ac{wb} condition $(124.86 \pm 26.53)$; a statistically significant increase of $37.54$ ($95\%$ CI, $30.30$ to $44.78$), $t(31)=10.58$, $p=.00$. We removed the data corresponding to participants who were stuck, and ran the same test again. The distances conformed to a normal distribution in both conditions as indicated by a Shapiro-Wilk test ($W(25)=0.944$, $p=.182$ for \ac{js} and $W(25)=0.954$, $p=.301$ for \ac{wb}) and there were no outliers. A paired samples t-test indicated that there was a statistically significant increase in the distances that the participants reached within a given time frame in \ac{js} condition from $132.83 \pm 23.45$ to $165.17 \pm 11.93$; an increase of $32.33$ ($95\%$ CI, $24.73$ to $39.94$), $t(24)=8.776$, $p=.00$.

\vspace{0.2em}
\noindent\textbf{Less collisions in \ac{js} condition (H2 rejected)}\hspace{0.5em} The number of crash events in \ac{js} condition $(m=0.4)$ were lower than the number of crash events in \ac{wb} condition $(m=1.4)$ for $N=32$, this decrease was statistically significant as indicated by a Wilcoxon Signed-Ranks test (two sided), $Z=-3.477$, $p=.001$, $r=0.615$. This tendency in having less collisions persisted also when the data corresponding to the people who got stuck in a collision-state for longer than $30s$ was removed $N=25$. There were significantly less crash events in \ac{js} condition $(m=0.4)$ as opposed to in \ac{wb} condition $(m=1.2)$, as indicated by a Wilcoxon Signed-Ranks test (two sided), $Z=-2.862$, $p=.004$, $r=0.572$.

\vspace{0.2em}
\noindent\textbf{Higher perceived workload in \ac{wb} condition (H3 rejected)}\hspace{0.5em} For each subscale of NASA-TLX, a Wilcoxon Signed-Ranks test (two sided) was performed to compare the ratings for \ac{js} and \ac{wb} conditions considering both the data with and without exclusions (see Table~\ref{tab:TLX_result} for respective means and the corresponding standardized test statistics, significance values, and effect sizes). With $N=32$ there was a statistically significant increase in the TLX scores in all dimensions other than performance. When the same test was performed with $N=25$, we observed that TLX scores were higher in all dimensions except performance and temporal demand. 

\begin{table*}
\centering
\begin{tabular}{|l|l|l|l|l|l|l|}
\hline
\multirow{2}{6em}{TLX subscale} & \multicolumn{2}{|c|}{Means (N=32)} & \multirow{2}{10em}{Test summary (N=32)} & \multicolumn{2}{|c|}{Means (N=25)} & \multirow{2}{10em}{Test summary (N=25)}\\
\cline{2-3}\cline{5-6}
& {\ac{js}} & {\ac{wb}} & & {\ac{js}} & {\ac{wb}} &\\ \hline
{Mental demand}             & 7.88      & 11.22          &    $Z=-3.73$, $p=.00$, $r=0.659$       &     7.28 & 10.56       & $Z=-3.47$, $p=.001$, $r=0.694$  \\ 
{Physical demand}           & 4.035     & 11.09          &    $Z=-4.63$, $p=.00$, $r=0.819$       &         4.96 & 6.48    &  $Z=-4.02$, $p=.00$, $r=0.804$  \\ 
{Temporal demand}           & 4.5       & 6.81           &  $Z=-2.64$, $p=.008$, $r=0.467$        &         4.08 & 10.28    & $Z=-1.82$, $p=.069$, $r=0.364$   \\ 
{Performance}               & 6.31      & 7.81           &    $Z=-1.40$, $p=.16$, $r=0.248$      &         6.44 & 7.60     &   $Z=-1.02$, $p=.307$, $r=0.204$ \\ 
{Effort}                    & 5.9       & 11.62          &    $Z=-4.55$, $p=.00$, $r=0.804$       &         5.6 & 11.12    &  $Z=-3.96$, $p=.00$, $r=0.792$  \\ 
{Frustration}               & 3.78      & 8.22           &     $Z=-4.38$, $p=.00$, $r=0.774$     &         3.48 & 7.88     &  $Z=-3.93$, $p=.00$, $r=0.786$  \\ 
\hline
\end{tabular}
\vspace{0.15em}
\caption{Means and the respective results of a Wilcoxon Signed-Ranks test (two-sided) for the \ac{tlx} subscales corresponding to $N=32$ and $N=25$ datasets.}
\label{tab:TLX_result}
\vspace{-0.4cm}
\end{table*}

\subsection{Exploratory results}\label{sec:exploratory_results}
We performed an exploratory analysis of all $32$ participants to get a deeper insight into our results.

\subsubsection{Quantitative Data}
%Easiness, confidence Likerts for sure. Presence forced-choice, maybe SUS - not sure if we want to report all the forced-choices

\vspace{0.2em}
\noindent\textbf{Participants felt more confident in using the \ac{js} condition}\hspace{0.5em}
To measure participants' confidence in using either control method, we asked them to rank their confidence in Likert-scale $(1-7)$ after each training session. People felt statistically significantly more confident in using \ac{js} $(Mean=6.25)$ as opposed to \ac{wb} condition, as indicated by a Wilcoxon Signed-Ranks test (two-sided), $Z=4.478$, $p=.00$, $r=0.79$.

\vspace{0.2em}
\noindent\textbf{\ac{js} condition was found easier to use}\hspace{0.5em}
We measured the relative ease-of-use by explicitly asking: Which control method was easier to use? $30$ out of $32$ participants $(94\%)$ found \ac{js} condition easier to use. This bias towards \ac{js} condition was statistically significant in an exact binomial test with exact Clopper-Pearson $95\%$ CI and had a $95\%$ CI of $79.2\%$ to $99.2\%$, $p=.00$ (two-sided).
In addition to the forced-choice question we also asked 7-point Likert-scale questions to measure ease-of-use. Comparing the ratings in \ac{js} condition $(Mean=6.03)$ with the ones in \ac{wb} condition $(Mean=3.59)$, we found that \ac{js} elicited a statistically significant increase in the comfort rankings, as indicated by a Wilcoxon Signed-Ranks test (two sided), $Z=4.783$, $p=.00$, $r=0.846$.

\vspace{0.2em}
\noindent\textbf{\ac{js} condition was preferred}\hspace{0.5em}
When asked explicitly which condition did the participants prefer, $26$ out of $32$ participants $(81\%)$ picked \ac{js} condition. An exact binomial test with exact Clopper-Pearson $95\%$ CI indicated that this bias towards \ac{js} condition was statistically significant and had a $95\%$ CI of $63.6\%$ to $92.8\%$, $p=.001$ (two-sided).

\vspace{0.2em}
\noindent\textbf{\ac{wb} condition did not increase presence}\hspace{0.5em}
When asked ``Thinking back to both of the experiences, which one gave a better sense of being in the robot's location?" $19$ participants $(59\%)$ picked the \ac{wb} condition, though this bias towards \ac{wb} condition was not statistically significant in an exact binomial test (two-sided), $p=.377$. %missing SUS here

\begin{figure}[]
    \centering
    \begin{subfigure}[b]{0.25\textwidth}
        \centering
        \includegraphics[width=0.99\textwidth]{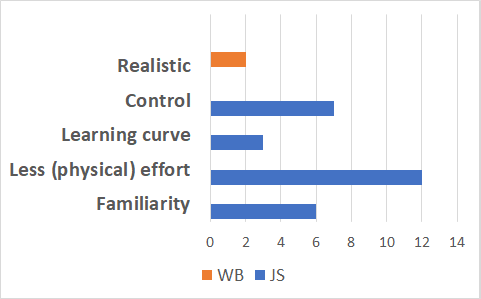}
        \caption{Ease-of-use}
        \label{fig:easy_codes}
    \end{subfigure}%
    \begin{subfigure}[b]{0.25\textwidth}
        \includegraphics[width=0.99\textwidth]{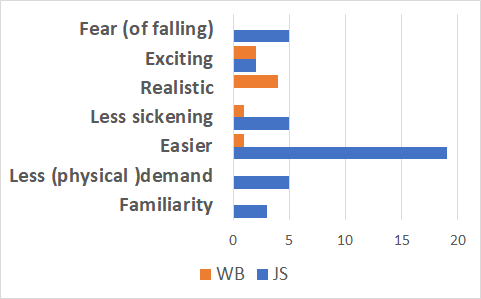}%\vspace{-2mm}
        \caption{Preference}
        \label{fig:pref_codes}
    \end{subfigure}%
\caption{Frequent codes found in open-ended questions regarding ease-of-use and preference together with their occurrences.}
\end{figure}

\subsubsection{Qualitative data}
The open-ended data was analyzed using the thematic analysis method with inductive approach \cite{patton2005qualitative}.
Fig.~\ref{fig:easy_codes} presents the frequent codes found in the data related to ease-of-use, divided by which condition was found easier to use. 
%The most common code was \emph{less physical effort} that is found in the $12$ responses given by the $30$ participants who found \ac{js} condition easier. This is followed by \emph{control} ($7$ occurrences) and \emph{familiarity} ($6$ occurrences) with the control method. Both responses given by the two participants who found \ac{wii} easier coded as the method being \emph{realistic}. 
\change{These codes are then grouped into three themes: expectations and previous experience, less (physical) effort, and control. The first theme refers to the previous experience of people in terms of moving in VR or in the real world and their expectations based on that. It encompasses the codes \emph{realistic}, \emph{learning curve}, and \emph{familiarity} containing 11 comments in total. Codes \emph{less (physical) effort} ($12$ occurrences) and \emph{control} ($7$ occurrences) are themes of their own referring to less required physical effort to use a control method and the sense of control over the robot, respectively. The majority of people who found \ac{js} condition easier thought so because it required \emph{less physical effort} (``It only required the movements of my thumbs, no more was required."). This is followed by reasons related to their \textit{expectations and previous experience} with VR. In particular, they felt \ac{js} was familiar and that it was easier to learn (coded as \emph{learning curve}). Finally, some people thought \ac{js} was easier to use because they had more \emph{control} over the robot with this method (``You feel more in control with this method"). Responses given by the two participants who found \ac{wb} easier stemmed from their \textit{expectations and previous experience}; they thought it was more \emph{realistic}, that is, similar to moving in the real world (``Personal choice of use by feet than hand, more realistic").}

\change{Codes and their frequencies related to the data on reasons for participants' preferences can be found in Fig.~\ref{fig:pref_codes}. The codes are grouped under the themes: \textit{bodily feelings} encompassing \emph{less physical demand} and \emph{less sickening} containing 11 comments, \textit{expectations and previous experience} encompassing \emph{realistic} and \emph{familiarity} containing 7 comments, \textit{emotions} encompassing \emph{fear} and \emph{exciting} containing 9 comments, and finally \emph{easier} being an important theme of its own. The most popular reason given by the participants who preferred \ac{js} condition was because it was \emph{easier} to use ($19$ occurrences in \ac{js} condition, for example, ``I didn't have to focus as much on how to use the control method on the first one...").
This was followed by the reasoning that fell under the theme bodily feelings, meaning that it induced less negative physiological response or that it was physically less demanding (``...The second one caused more motion sickness-like symptoms."). Some people who preferred \ac{js} thought that it made them feel better emotionally (falling under the theme \emph{emotions}), compared to \ac{wb}, especially since they did not experience a fear of falling down (``I didn't feel like falling"). People who preferred the \ac{wb} condition named the biggest factor as it being \emph{realistic} based on their \textit{expectations and previous experience} (for example, \textit{"Despite being more physically challenging it gives more movement and feels almost like real movement."}). Other reasons stated fell under having better \emph{emotions} (\textit{``It was cool and made me feel more in the place."}) and having less bad \emph{bodily feelings} (\textit{``And I also felt really nauseous doing the task with the controllers but not nauseous at all doing it with the balance board"})  .}

\section{Discussion}

\subsection{Original hypotheses}
On average, \ac{js} condition induced less VR sickness measured using the SSQ total score. Thus, refusing H1.
The answers to open-ended questions reveal interesting insights and possible reasons for this result on VR sickness.
%there was no difference in VR sickness through SSQ, rejecting H1. 
Moreover, in the open-ended questions the \emph{less sickening} code was used as a reason by 5 participants who preferred the \ac{js} and only once by who preferred the \ac{wb}, hinting towards participants suffering more from VR sickness while using the board than with the joystick. Similarly, H2 and H3 were rejected too; participants reached farther along the path while using the joystick as opposed to the \ac{wii}, and the \ac{tlx} scores showed that the participants perceived controlling the robot using joysticks less demanding.
%participants considered joystick easier. 

Further examination of the answers to the open-ended questions reveals a potential relation between perceived difficulty and VR sickness.
%seem connected with each other. 
Two of the participants who preferred the \ac{js} condition because it induced less sickness stated that \textit{``Not having to look at floor while leaning prevents nausea"} and \textit{``I didn't have to focus as much on how to use the control method on the first one. The second one caused more motion sickness-like symptoms."} Since staring at the floor would result in more optical flow, and thus likely induce more VR sickness, if participants looked at the floor more with the Wiiboard, it may likely have increased their VR sickness levels.

%In relation to this, we note that even though we carefully told participants to follow the general direction of the line, and not stay exactly on top of it, several participants still tried to stay directly on top of the line. 

A potential relation between difficulty and VR sickness was further indicated 
%The chance of difficulty causing more VR sickness is further enhanced
by looking at the confidence scores of the participants who commented their respective method to be less sickening. The participant who found the Wiiboard less sickening gave a confidence score of 7 (on a 7-point Likert-scale) for using the \ac{wb}, whereas the average of the confidence scores given by the 5 people who found \ac{wb} more sickening was $5.2$. 
Similarly, one participant who mentioned that the \ac{wb} condition was less sickening rated its ease-of-use as $5$ (out of 7), whereas the average of the ease-of-use ratings given by the 5 participants who found the board more sickening was $2.6$. 
%got less sick considered the easiness of using the board 5, whereas the average of the five participants considering the board more sickening was $2.6$. 
Despite being based on a small sample size, these insights further strengthen the implication of difficulty causing more VR sickness. The other possible cause for increased VR sickness with the board is the postural instability caused by the control motions, even though, as mentioned, the order of cause and consequence is unclear \cite{litleskare2021relationship} and not researched when leaning is used for control.

Interestingly, earlier papers \cite{de2008using} described the board feeling ``easy" when used among the researchers and colleagues. Similarly, all our pre-pilot users did prefer the board and effectively used it to control the robot. However, during the study, regardless of the unlimited training time, the participants still perceived using board as more difficult ($Mean=3.59$ on a 7-point Likert-scale). We expected that participants would find using the board more difficult compared to joystick; this was in part due to the well known result in 3D user interfaces such that that using a high \ac{dof} input to control a 
%lower number of inputs 
lower \ac{dof} system
is typically challenging \cite{laviola20173d}, and in part due to the familiarity of the general population with the joystick. 
However, we did not expect to see a difference in the perceived difficulty of using the \ac{wii} to the extent that we observed
%but not to the extent we perceived 
(statistically significant difference in favor of \ac{js} in almost all dimensions of the \ac{tlx}). Besides VR sickness, the difficulty had a major impact on %the choice of 
preference: 20 out of 32 participants, stated "easiness" as their reason for preferring a particular method,
%choosing the particular method as preferred, 
among which 1 person stated it in preference of the Wiiboard and 19 in favor of the joystick. Additionally, when asked why one method felt easier than the other, three people specifically noted the steep learning curve for the Wiiboard \textit{(``The second method is a bit harder to get used to")} and six people mentioned that their familiarity with joysticks may have helped
%considered the familiarity of the joystick to have helped
\textit{(``Joysticks are a familiar method and very intuitive too. Balancing takes more effort")}.

%\textit{(``First was more fun and demanding, second was easier. In the first one I had more symptoms of nausea and dizziness, so that is why I choose the second.")}

\subsection{Exploratory data}
Chen et al.~\cite{chen20136dof} found, in an old study, that stepping-based (head-tracked, with steps taken at each direction to move towards that direction) navigation caused less VR sickness and induced a (moderately) greater feeling of presence than a joystick. However, there are several differences between their study and ours.
%However, there's several differences between the studies; 
First, Chen et al.~considered 6D-navigation without physical limits. Moreover, all of their 20 participants were experienced with VR, and it was not explained how much prior experience they had with the system; also, they used a different kind of joysticks. 
Thus, even though there is evidence that self-rotation, 
easily achieved by the stepping method,
%easier allowed with the stepping method, 
reduces VR sickness and increases the feeling of presence and spatial awareness \cite{nguyen2019naviboard}, it is not clear which method in general would be more useful for telepresence robot navigation. A method combining leaning and self-rotations could probably provide best of both worlds.

%There were also interesting open-ended answers regarding the physicality of the control and fear of falling; 13 out of 32 participants considered less physical effort to be a major reason why one method was easier. Additionally, such body-based controls requiring standing are not very inclusive to non-healthy users. Thus, also joystic-based navigation does have it's place. 

We decided to make participants undergo a training session
without the HMD. 
We wanted to give the participants as much time as they needed and 
did not want to limit the training duration because we expected that they would not be familiar with the board and that using the board would employ a steeper learning curve. 
%(and non-equal practice times with the HMD on would have corrupted the SSQ scores) 
Our expectations were supported by the participants' comments;
for example, \textit{``More exciting and would ``become" easier after many practices.", ``I'm used to moving in VR with joysticks, the balance board was a bit harder to get used to especially because the HMD made it a bit harder to balance myself."}. However, a longer training with the HMD on could have caused excessive sickness and potential differences in training times could have corrupted the \ac{ssq} data. Despite mitigating these potential issues, practicing without the HMD caused a shortcoming in our approach, mentioned also in the comment above, that is, using \ac{wb} with the HMD was slightly different than using it without. It would be interesting to perform a smaller-sample qualitative long-term study with the board, where we give participants ample time to practice on the board on the scale of days or weeks, and then compare the differences and SSQ scores; we suspect that this kind of study would better bring out the strengths of the leaning-based control using a balance board.

We also wondered whether the body-based locomotion would make participants feel more present; whereas it is well established that having an actual virtual body that tracks your motions increases presence \cite{slater2009place}, there is, to the knowledge of the authors, no clear evidence whether having ``a bit more realistic" body-based control should increase the feeling of presence (in \cite{slater1998influence} simply moving the body increased presence, but the comparison was standing still, not moving via a joystick or similar). There were hints towards the board making participants feel more present; four participants preferred the Wiiboard because it felt more \emph{realistic} \textit{("It was cool and made me feel more in the place. Although it was easier with the first method and I think that I also accomplished the task better with the first method (like no bumping into the wall).")}, and some participants who preferred joystick at the end still stated the positives of the Wiiboard towards that direction \textit{(``Because I could accomplish the task with the first one and with the second I got stuck. However, the second one was more fun and I felt more engaged.")}. However, the forced-choice question about presence was not significant (even if leaning towards that direction), and the SUS showed no significant difference either. Nonetheless, there seems potential for more studies on presence with body-based locomotion.

\subsection{Limitations and future work}
A clear limitation of the study was the hardware: the Wiiboard is an old technology and has limited sensors, which are not extremely accurate. This could have been a reason for nine people mentioning better control of the robot with the joystick as a reason for finding it easier. Also, altogether 5 participants reported fear of falling off the board as a reason for preferring the joystick; it seems that a thin mattress, or carpet, with sensors would be better to alleviate this fear. Additionally, such a system would allow rotating the robot also via rotating the whole body, which has been shown useful in reducing VR sickness and enhancing spatial awareness \cite{nguyen2019naviboard}. Such a combination of the strengths of the methods could provide interesting results. 

These results could also be used as a stepping stone to find out the relationship between postural instability and VR sickness; does active destabilization of the posture, in the form of leaning, contribute to VR sickness? With a more accurate pressure mat, we may be able to differentiate between wanted and unwanted postural sway, and correlate them with the (perhaps continuously measured) VR sickness and HMD tracking data; additionally, if the pressure sensor was used to measure the postural instability also when controlling via the joystick, this comparison could reveal interesting facets about wanted and unwanted postural instability. 

\section{Conclusion}\label{sec:con}
The study presented in this paper reveals that leaning-based methods on a balance board have potential, but the unfamiliarity and difficulty of the Wiiboard prevented the possible positive effects of reducing VR sickness. We identified several reasons making the board so difficult, such as not enough responsiveness, steep learning curve requiring more training than a few minutes before the actual study, training with the headset on even with the chance of inducing VR sickness during training, and the height of the board causing a fear of falling. A more responsive and larger or flat board, which could also allow self-rotations, could be tested for chances of improvement. 

%% if specified like this the section will be committed in review mode
\acknowledgments{This work was supported by a European Research Council Advanced Grant (ERC AdG, ILLUSIVE: Foundations of Perception Engineering, 101020977), Academy of Finland (projects PERCEPT 322637, CHiMP 342556), and Business Finland (project HUMOR 3656/31/2019).}

\bibliographystyle{abbrv-doi}

\bibliography{references}
\end{document}